\documentclass[letterpaper, 10 pt, conference]{ieeeconf}  % Comment this line out if you need a4paper

\IEEEoverridecommandlockouts                              % This command is only needed if 
\usepackage{amsmath} % assumes amsmath package installed
\usepackage{graphicx}
\usepackage{comment}
\usepackage{subfig} % qilu 01/29/2024 make it possible to include more than one captioned figure/table in a single float
\usepackage{multirow} % for tables
\usepackage{xcolor}
\usepackage{algorithm}
\usepackage{cite}
\usepackage{url} 
\usepackage[noend]{algpseudocode}

\title{\LARGE \bf
Non-Contact Vibration-Based Damage Detection of Civil Structures Using a Cost-Effective Autonomous UAV
}

\author{Javier Becerril$^1$, Maximiliano Vargas$^2$, Jennifer Herrera$^3$, Joanna Gutierrez$^1$, Jorge Rios$^1$, \\ Mohsen Amjadian$^3$, Constantine Tarawneh$^2$, Jinghao Yang$^4$, and Qi Lu$^{1,*}$ % <-this % stops a space
\thanks{$^1$Department of Computer Science, The University of Texas at Rio Grande Valley (UTRGV), Edinburg, TX, USA.}
\thanks{\tt\small \{javier.becerril01, joanna.gutierrez03, jorge.rios03, qi.lu\}@utrgv.edu }
\thanks{\raggedright $^{2}$Department of Mechanical Engineering at UTRGV. \tt\small \{maximiliano.vargas01, constantine.tarawneh\}@utrgv.edu} 
\thanks{\raggedright $^{3}$Department of Civil Engineering at UTRGV. \tt\small \{jennifer.herrera07, mohsen.amjadian\}@utrgv.edu} 
\thanks{$^{4}$Department of Electrical and Computer Engineering at UTRGV.} 
\thanks{\tt\small jinghao.yang@utrgv.edu}
\thanks{$^*$Corresponding author}% <-this % stops a space
}

\begin{document}

\maketitle
\thispagestyle{empty}
\pagestyle{empty}

%%%%%%%%%%%%%%%%%%%%%%%%%%%%%%%%%%%%%%%%%%%%%%%%%%%%%%%%%%%%%%%%%%%%%%%%%%%%%%%%
\begin{abstract}
This paper presents a non-contact approach for vibration-based structural damage detection using an autonomous and customized cost-effective unmanned aerial vehicle (UAV). Vibration signals are extracted from video recordings through vision-based motion tracking to identify shifts in natural frequencies indicative of structural degradation. A laboratory-scale frame structure is evaluated under healthy and simulated-damage conditions, where damage is introduced via an added mass. The proposed system is validated through a multi-platform experimental study involving two high-resolution smartphones, a USB camera, and a custom-built low-cost UAV equipped with an onboard camera and an AprilTag-based autonomous alignment system for operation in GPS-denied environments. The displacement time is extracted and analyzed in the frequency domain and compared to reference measurements from contact accelerometers and a finite element model. 
Experimental results show that all platforms successfully capture the fundamental frequency and its shift due to damage. Although the UAV exhibits slightly higher errors (approximately $5–6\%$) due to platform-induced disturbances and sensing limitations, it reliably detects damage-induced frequency changes. Compared to commercial UAV systems, the proposed platform achieves comparable inspection performance at significantly lower cost. These results demonstrate that low-cost autonomous UAVs provide a practical, flexible, and scalable solution for structural health monitoring, particularly in scenarios where contact-based sensing is impractical. The findings also support the potential for the deployment of multiple cooperative UAVs to further enhance inspection coverage and robustness.

\end{abstract}

%\begin{IEEEkeywords}
%Swarm Robotics, Intrusion Detection, Foraging Robots, Pheromone Trails, Security in Robotics
%\end{IEEEkeywords}

\section{Introduction}
\label{Introduction}

% from Jennifer
%\subsection{Background}
Reliable and cost-effective vibration measurement remains a central challenge in structural health monitoring (SHM)~\cite{Limongelli2019, Brownjohn2011}. Conventional contact sensors, such as accelerometers, provide accurate dynamic response data but require physical installation, wiring, and maintenance, which can limit scalability for large or difficult-to-access civil structures like long-span bridges~\cite{Wong2007}, the underside of bridge decks that require scaffolding or rope access, live electrical transmission towers, and contaminated areas where contact is a safety concern. Non-contact optical techniques based on computer vision have therefore gained increasing attention as flexible alternatives~\cite{ABEDIN20214012}.

Beyond fixed stand-off camera systems, unmanned aerial vehicles (UAVs) offer additional flexibility for non-contact remote structural monitoring. UAV-mounted cameras can access elevated or hazardous locations without requiring permanent sensor installation. However, integrating UAV-based imaging with vibration-based damage detection requires validation against control experimental conditions to ensure accuracy and robustness. In this study, we experimentally evaluate vibration-based damage detection using a combination of stand-off cameras and a low-cost UAV. A one-story frame structure is tested under 2 excitation conditions: (1) control harmonic loading using an electrodynamic shaker and (2) free-vibration excitation for UAV-based measurements. Dynamic responses are captured using high-resolution smartphones, a USB camera with higher frame rates, and a drone-mounted camera system. Motion tracking is performed using Motion Tracker Beta~\cite{FLOCH2023101424}, an open-source image-based tracking tool, to extract displacement time histories from recorded videos. The frequency-domain analysis is then conducted to identify the fundamental frequency and each shift under simulated damaged conditions. Experimental results are compared with accelerometer measurements and validated against the finite element (FE) model developed in COMSOL Multiphysics.

Compared to previous UAV-based structural monitoring work~\cite{ReliabilityUAV2023,UASBridgeTwoCameras2024}, this paper presents a low-cost custom UAV with an autonomous AprilTag-based alignment system that positions and maintains camera orientation in a GPS-denied indoor environment without manual piloting, addressing an operational gap in~\cite{ReliabilityUAV2023} in which UAVs must be manually positioned before data collection. We also performed a multi-platform experimental evaluation comparing two smartphones, a USB camera, and the custom UAV on the same structure, demonstrating comparable damage detection at significantly lower cost than the commercial systems used in our previous work~\cite{amjadian2026dic}. By assessing the trade-offs between spatial resolution, temporal accuracy, and platform flexibility, this work demonstrates that a low-cost, custom-developed UAV with an ordinary camera system can achieve a similar performance as advanced commercial UAVs for practical vibration-based structural health monitoring. Finally, we show that the low-cost UAV successfully identifies damage-induced frequency shifts consistent with contact sensor measurements, validating its feasibility for autonomous SHM in GPS-denied environments.

%Recent innovations, such as the integration of compressed sensing with DIC, further expand the potential of the DIC method. Kato and Watahiki~\cite{KATO2023110495} introduced randomized single-exposure sampling (RSES) using low-speed cameras and strobe lighting to identify high-frequency vibration modes. Meanwhile, Wang et al.~\cite{WANG2023113041} investigated the effects of image compression, demonstrating that frequency accuracy can be preserved even with reduced storage demands. 

\section{Related Works} %dont change it to subsection
DIC and feature-based motion tracking are non-contact full-field optical measurement techniques used to quantify surface displacements and strains by tracking image patterns or distinct features in sequential images~\cite{KATO2023110495,WANG2023113041}. In large-scale civil structures, Malesa et al.~\cite{Malesa2010} demonstrated full-field displacement monitoring on a railway bridge as trains passed, showing the feasibility of field-scale applications. Studies by VanDyk and Simha~\cite{VanDykSimha2024} and Lin et al.~\cite{LinDIC2023} have demonstrated that reliable full-field dynamic characterization can be achieved using low-cost consumer-grade digital cameras, simplified single-camera setups, and open-source software for image processing and data analysis. This makes DIC especially appealing for SHM applications in resource-constrained settings or for large-scale infrastructure systems where the deployment of dense sensor networks is impractical. 

Recent work in~\cite{CrossCorrUAV2017} has established the feasibility of extracting natural frequencies from UAV videos, where consumer-grade cameras on a UAV were used to conduct system identification. In~\cite{ChenUAVDIC2021}, it combined the use of UAV video with DIC to extract vibration measurement of a bridge model by analyzing the videos to track the displacement of the measurement points. In~\cite{ReliabilityUAV2023}, a study was conducted on UAV vision-based displacement measurement systems to analyze actual structure measurements. In~\cite{Ri2024DronebasedDM}, a drone-based displacement measurement framework was introduced for on-site bridge girder inspections. In~\cite{UASBridgeTwoCameras2024}, it inspects bridge displacement using two cameras mounted on a UAV. One camera is focused on taking measurements near the midspan, while another one is used to stabilize the UAV using stationary locations at the bridge supports. However, the cost of commercial UAVs is high and cannot be programmed to be autonomous. In addition, the use of AprilTag localization, which improves the estimation of the pose of the UAV in an indoor environment, as studied in~\cite{UAVAprilTags2020}. The use of optimized optical flow sensors has been validated as an effective approach to stabilization and positioning of UAVs in indoor and GPS-denied environments~\cite{VibrationOpticalUAV2025}. In their study, they examined the reliability of the dynamic measurement of the system by measuring various vibrations with varying frequencies applied through free vibration excitation. They compared the results obtained from the UAV to those of conventional displacement sensors and concluded that the UAV system demonstrated results similar to those of conventional methods. 

\section{Methodology}
\label{method}

\subsection{Vision-Based Motion Tracking}
Vision-based motion tracking is a non-contact optical technique used to measure structural displacements by analyzing image sequences captured during motion. In this approach, the movement of identifiable surface features between a reference image and subsequent frames is tracked to estimate displacement over time. Correlation-based tracking methods, such as those employed in DIC~\cite{SchreierOrteuSutton2009}, determine motion by comparing image subsets between undeformed and deformed states. However, unlike full-field DIC, which is primarily designed to compute displacement and strain fields over an entire surface, the present study focuses on tracking selected points of interest to obtain displacement time histories for vibration analysis. In practice, a high-contrast surface pattern (natural texture or artificial speckle) improves tracking robustness by providing distinguishable features. The recorded images are divided into small regions, and a correlation or feature-matching algorithm is used to determine the relative shift of these regions between frames. Sub-pixel interpolation techniques are applied to enhance the measurement resolution.

The underlying principle assumes brightness constancy, meaning that the gray-level intensity of a material point remains unchanged during motion. This assumption can be expressed as follows.

\begin{equation}
f(x,y) = g\big(x + u(x,y),\, y + v(x,y)\big),
\end{equation}

where $f(x,y)$ and $g(x,y)$ denote the gray-level intensities in the reference and subsequent images, respectively, and $u(x,y)$ and $v(x,y)$ represent the displacement components in the horizontal ($x$) and vertical ($y$) directions.

In this study, a single-camera 2D vision system is used to track structural vibrations. The extracted displacement time histories are analyzed in the frequency domain to identify dynamic characteristics such as natural frequencies and their shifts due to structural damage. This motion-tracking-based approach provides a non-contact, cost-effective, and easily deployable solution for structural health monitoring applications.

\subsection{Frame Building Prototype Design and Fabrication}

\begin{figure}[htbp!]
    \centering
    \includegraphics[width=3.0in]{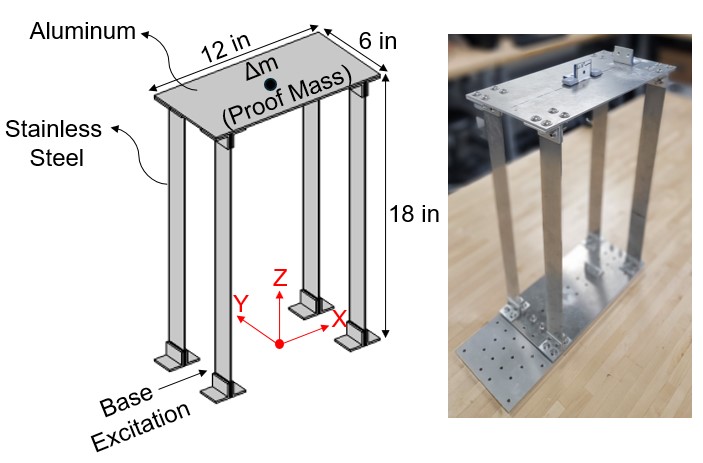}
    \vspace{-2mm}
    \caption{Prototype of the scaled one-story frame building fabricated for vibration testing.}
    \label{fig_prototype}
\end{figure}

To experimentally evaluate the performance of the non-contact vibration measurement method, a small-scale one-story frame prototype is designed and fabricated in our Structural Engineering and Materials (STEM) Lab. As shown in Fig.~\ref{fig_prototype}, the prototype represents a simplified building structure consisting of four stainless-steel columns, each with a height of 18 inches, a width of 1.5 inches, and a thickness of 1/16 inch, supporting a rigid aluminum roof diaphragm with dimensions of 12 inches in length, 6 inches in width, and $1/8$ inches in thickness. The columns, at their top ends, are connected to the roof diaphragm, and at their bottom ends, are anchored to a rigid base plate on the shaker’s slip table, both with brackets and bolts, ensuring stable excitation during harmonic loading tests. The geometric scale was chosen to remain compatible with the capacity of the vibration shaker while preserving dynamic characteristics representative of a low-rise flexible building. 

A 3D FE model of the frame is developed in COMSOL Multiphysics v6.2 to validate the experimental results. The FE model was meshed with free tetrahedral elements, with a maximum element size of 45 mm and a minimum of 8 mm. The columns are modeled as stainless steel with an elastic modulus of $200$ GPa, A Poisson's ratio of $0.3$, and a mass density of $7850$ $kg/m^3$, while the roof diaphragm is modeled as aluminum with an elastic modulus of $70$ GPa, a Poisson’s ratio of $0.33$, and a density of $2700$ $kg/m^3$. The bottom ends of the columns were fixed at the base to replicate the bolted boundary condition on the slip table. %Fig.~\ref{fig_prototype}

To simulate excitation, the frame was subjected to a Linear Chirp Signal (LCS) applied along the $x$-axis, with the base acceleration defined as

\begin{equation}
A_b(t) = A_{b0}\sin[2\pi(f_a + \frac{f_b - f_a}{2T}t)t]
\end{equation}

where $A_{b0}$ is the maximum acceleration of the ground, $f_a$ and $f_b$ are the initial and final excitation frequencies, and $T$ is the duration of the sweep frequency from $f_a$ to $f_b$. This FE model serves as a reference for identifying the natural frequencies and comparing vibration responses obtained from both contact-based (i.e., accelerometers) and computer vision-based non-contact measurements using stand-off cameras and UAV. The FE model represents only the undamaged structural condition. 
\begin{comment}

I added this part "The FE model represents only the undamaged structural condition. "
\end{comment}  

\subsection{UAV Design and Components}

%From Javier
\subsubsection{Hardware Development} 
\begin{figure}[htbp!]
    \centering
    \includegraphics[width=2.8in]{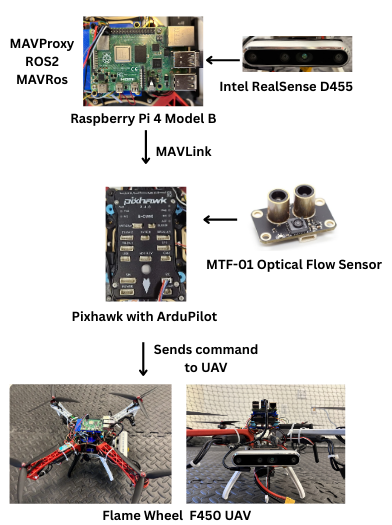}
    \vspace{-2mm}
    \caption{System architecture and workflow.}
    \label{fig_UAV_Visual_Pose}
\end{figure}

The proposed quadrotor UAV platform includes a DJI FlameWheel 450 frame, a Pixhawk flight controller running the ArduPilot firmware (chosen for its straightforward compatibility and documentation; PX4 can also be used), and a Raspberry Pi 4 Model B companion computer responsible for perception and high-level control. The flight stabilization is handled by the hierarchical flight controller, while visual perception and relative pose estimation are executed on the companion processor (see Fig.~\ref{fig_UAV_Visual_Pose}). 

The flight controller serves as the stabilization unit using the onboard Inertial Measurement Unit (IMU), composed of accelerometers and gyroscopes, to estimate vehicle attitude and angular rates. These measurements are processed internally by ArduPilot’s control loops to generate motor commands that regulate roll, pitch, yaw rate, and thrust.

A downward-facing MTF01 optical flow sensor is connected directly to the flight controller and is configured to provide velocity measurements relative to the ground plane. When operating in GPS-denied mode, these optical flow data help onboard controls reduce horizontal drift (see Fig.~\ref{fig_UAV_Visual_Pose}). The MTF01 module also incorporates an onboard rangefinder that measures the distance to the ground. The optical flow sensor will provide positioning on the $x$ and $y$ axes, while the rangefinder measures the position of the $z$ axis. Together, these sensors enable drift-reduced hovering and localization without requiring global positioning signals.

A forward-facing camera module (Intel RealSense Depth Camera D455) is mounted on the frame and connected to the companion computer that runs a ROS-based perception pipeline. In experimental trials, the camera is used to acquire the vibration signal. The UAV can detect AprilTags~\cite{AprilTag2011} in the frame building and compute the relative pose. Once the UAV detects the Apriltags, it tracks the Apriltags automatically, and the camera always faces the frame building for data acquisition (see Fig.~\ref{fig_phy_UAV_camera_view}).

The total hardware cost of our UAV is approximately \$1,000 at the time of assembly, including the DJI FlameWheel 450 frame (\$400), the MTF01 optical flow sensor (\$30), the Intel RealSense D455 camera (\$450), and the Raspberry Pi 4 (\$60). These components were purchased individually at retail price, but if purchased in bulk or scaled in production, the cost would be reduced. For comparison, commercially available consumer drones, such as in~\cite{ReliabilityUAV2023}, the DJI Mavic 3 series range from approximately \$2,000 to \$2,500, while higher-end platforms such as the DJI Matrice 210 in~\cite{UASBridgeTwoCameras2024} retail around \$10,000. Although commercial systems provide integrated stabilization and imaging hardware, they lack the level of extensibility and modularity required for research-oriented experimentation. Therefore, the proposed custom platform offers a cost-effective and scalable alternative that enables flexible structural health monitoring research. In addition, the UAV platform can be easily upgraded and adapted for deployment in other application domains, further enhancing its versatility and long-term utility.

\subsubsection{Software Systems} The system is conceptually organized into three hierarchical layers: stabilization, perception, and command. The stabilization layer is implemented entirely onboard the flight controller and is responsible for low-level state estimation and control. It performs attitude estimation and regulation using measurements from the onboard inertial measurement unit (IMU). In addition, optical-flow-based velocity aiding is integrated to mitigate horizontal drift in GPS-denied and indoor laboratory environments. All motor commands are generated exclusively by the stabilization layer, which computes the required control inputs and maps them to individual motor outputs.

The perception layer runs on the companion computer using ROS. This layer processes camera imagery, detects AprilTags, and computes relative pose measurements between the UAV and the visual reference. During operation, these pose estimates are used to autonomously align the UAV with the target structure, adjusting position, camera orientation, and stand-off distance until the AprilTag is centered in the camera's field of view. Once alignment is achieved, the system transitions to a hover hold mode in which optical flow velocity estimation and IMU-based attitude stabilization take over position maintenance. Continuous AprilTag-based corrections are deliberately not applied during data collection, as fine-scale real-time corrections would cause constant UAV micro-adjustments, introducing additional motion noise into the recorded signal. The UAV, therefore, maintains position within a tolerance band during recording.

\subsubsection{Simulation} 
\begin{figure}[thpb]
\centering
\subfloat[]{
\label{fig_simulated_UAV}
  \includegraphics[width=0.21\textwidth]{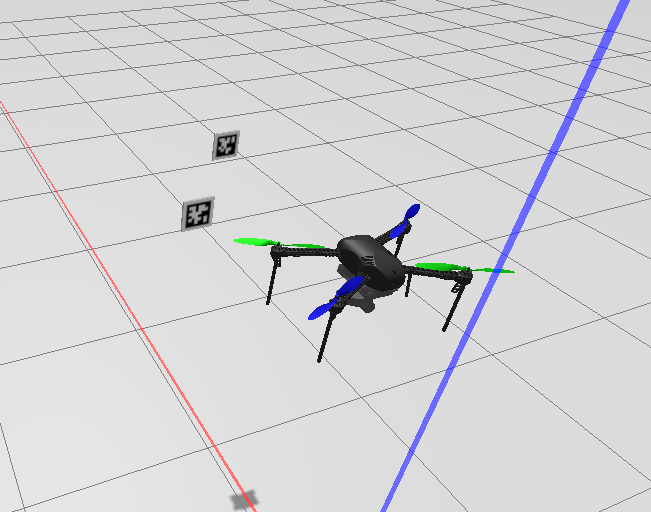}
}
%\vspace{1mm}
\subfloat[]{1
\label{fig_phy_UAV_camera_view}
  \includegraphics[width=0.25\textwidth]{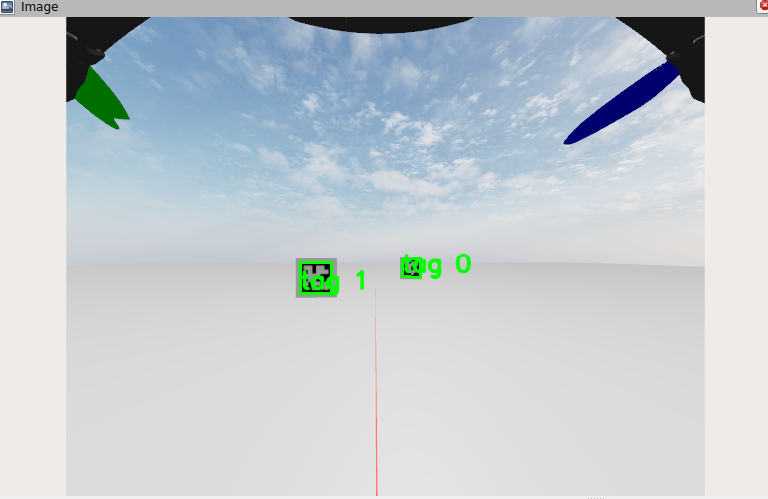}
}
\caption{The simulation of the UAV. (a) Developed UAV and two AprilTags in Gazebo Simulator; (b) Two AprilTags are detected, and the image is from the UAV's camera view.}
\label{fig_sim_UAV_camera_view}
\end{figure}

The simulation environment replicates the physical architecture. Validates the UAV's design efficiently before we run the physical UAV (see Fig.~\ref{fig_simulated_UAV}). In simulation, the UAV model uses an IMU-driven stabilization loop comparable to the Pixhawk’s onboard controller. AprilTag detection is performed using camera input, and the resulting relative pose measurements are actively used within the control loop (see Fig.~\ref{fig_sim_UAV_camera_view}). The simulated vehicle is capable of locking onto the visual marker, regulating lateral position error, and yaw alignment with respect to the tag. This enables evaluation of vision-referenced hover stabilization and relative pose control.

The AprilTag detection and pose estimation pipeline is operational at the perception level. The stabilization of the UAV during experimental trials is based on IMU-based attitude control and optical-flow-assisted velocity estimation. %Integration of real-time visual pose feedback into the control architecture is in active development.

The simulation framework serves as a validation platform for the intended vision-referenced control strategy, while the current hardware experiments evaluate the sensing stack and drift characteristics independently. Future work will focus on incorporating AprilTag-based relative pose measurements into the control architecture to enable vision-referenced hover and drift correction. In this planned extension, the companion computer will compute the relative position and yaw error with respect to a visual marker and use these errors to generate velocity or position setpoints. These setpoints will then be transmitted to the flight controller via MAVLink in Guided or Guided-NoGPS mode, allowing the vision system to provide visual reference corrections while the flight controller continues to maintain attitude stabilization.
%end for Javier

\section{Experimental Setup}
\label{experiment}

In this study, we had two experimental setups. The first is a forced-vibration experiment by a shaker excitation for smartphones and USB cameras. The second is a free-vibration experiment by a frame structure without a shaker excitation for the UAV. Due to flight safety constraints and logistical limitations, the shaker could not operate within the UAV flight test enclosure. However, both experiments theoretically lead to the same result, namely the frequency of the scaled frame, which remains consistent in both free and forced vibration cases, regardless of whether mass is added or not.

% from Jennifer
\subsection{Forced-Vibration Testing} 

A small-scale, one-story structural frame was designed and fabricated to facilitate laboratory vibration testing. The frame consisted of four stainless-steel columns that supported a rigid aluminum roof diaphragm. The prototype was mounted on a vibration shaker, which applied controlled harmonic excitation. In addition, free vibration tests were performed using a drone-mounted camera, which captured natural responses without external excitation.

For the smartphone and USB camera tests, we recorded videos of the frame from a fixed distance, ensuring consistent lighting and camera placement. Both devices captured images at 30 and 60 frames per second. The displacement data was extracted by manually tracking a point on the roof diaphragm in each video. After converting pixel displacements to physical units using calibration, the natural frequencies were derived by analyzing the displacement time histories in the frequency domain.

\begin{figure}[htbp!]
    \centering
    \includegraphics[width=2.7in]{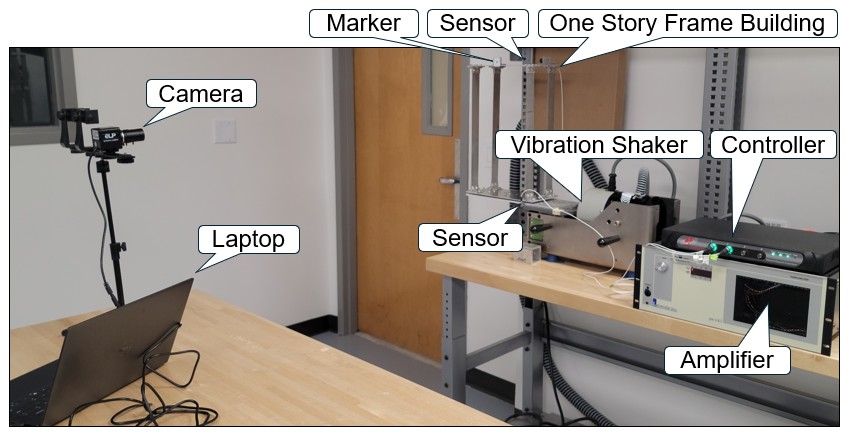}
    \vspace{-2mm}
    \caption{Forced-Vibration experimental setup for testing the non-contact camera methods.}
    \label{fig_camera}
\end{figure}

The experimental setup is shown in Fig~\ref{fig_camera}. It consists of the prototype of the scaled frame mounted on the slip table of an electrodynamic vibration shaker. The shaker is operated using a dedicated controller, which also serves as a data acquisition (DAQ) system for recording the responses of conventional contact sensors. To provide reference measurements, two PCB accelerometers are installed, one on the rigid aluminum diaphragm at the roof level and another on the base plate of the frame. The shaker is driven through a power amplifier that regulates the input voltage and ensures stable harmonic excitation in the longitudinal direction of the frame. 

\begin{figure}[htbp!]
    \centering
    \includegraphics[width=2.6in]{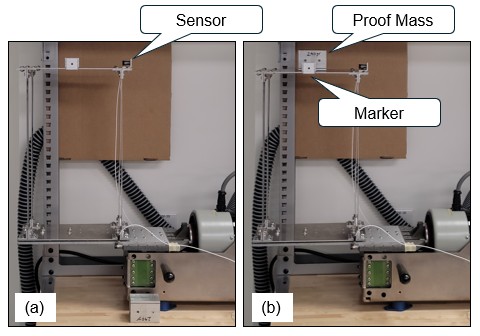}
    \vspace{-2mm}
    \caption{Frame building testing: (a) healthy and (b) damaged.}
    \label{fig_close_view_frame}
\end{figure}

Three cameras were used to capture the dynamic response of the structure during the force vibration testing: (1) a Samsung Galaxy S21 Ultra 5G and (2) an iPhone 15 Pro, both offering high spatial resolution but limited frame rates, and (3) an IEights USB camera that provides lower resolution with a higher frame rate. The cameras are positioned at a stand-off distance of approximately 4 ft to capture the motion of a selected point of interest on the rigid diaphragm and record displacement time histories for DIC and motion-tracking analysis. This point of interest is marked with a large black circle on a white background and attached to the rigid diaphragm, as shown in Fig.~\ref{fig_close_view_frame}.

\subsection{Free Vibration Testing} 

In contrast, the UAV measurements were collected by hovering the drone at a fixed distance from the frame during free vibration tests. Each video was processed using a motion tracking algorithm, which automatically tracked the displacement of the same reference point. The frequency analysis from the UAV data was then compared directly to the results of the smartphone and the USB camera. Using both approaches, we assessed how well a low-cost UAV system could capture structural vibrations compared to traditional smartphone and USB recordings.
% end 

\begin{figure}[thpb]
\centering
\subfloat[]{
\label{fig_UAV_Vibration_Collection}
  \includegraphics[width=0.21\textwidth]{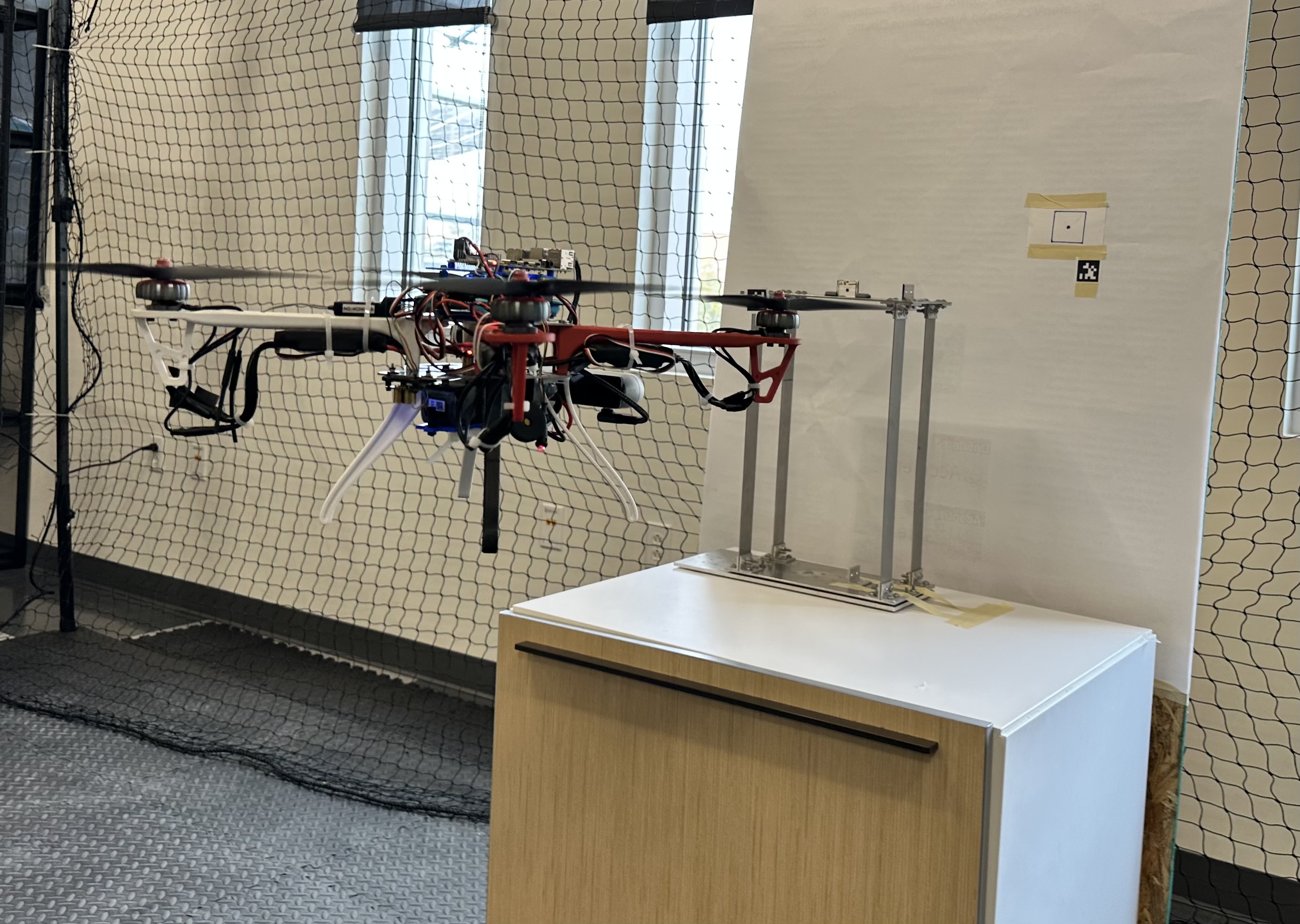}
}
%\vspace{1mm}
\subfloat[]{
\label{fig_phy_UAV_camera_view}
  \includegraphics[width=0.25\textwidth]{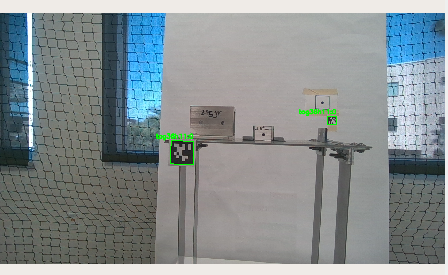}
}
\caption{Experimental setup. (a) The UAV is collecting data; (b) The UAV's camera detects two AprilTags.}
\label{fig_phy_UAV}
\end{figure}

Fig.~\ref{fig_UAV_Vibration_Collection} shows the fabricated frame building set up in our MARS lab (Multiple Autonomous Robot Systems) for the free vibration testing with point and ArUco markers. The UAV is recording the free vibration signal in the cage of the flight test field of the UAV. If operators manually control it, it is difficult for the UAV to focus on the marker for more than a minute, since the UAV has a minor drift in localization, which is very common in an indoor environment without GPS signals. The localization of the UAV is affected by noise in the laboratory environment (e.g., magnetic fields from equipment and facilities). 

In this experiment, the UAV was programmed to autonomously detect and track an AprilTag attached to the target frame. We assume that the UAV is deployed within a region where the marker is detectable after takeoff. To account for platform-induced disturbances, an additional AprilTag was mounted on a nearby stationary board as a reference. The UAV inherently generates vibrations from its motors, which are superimposed on the measured vibrations of the frame. However, these platform-induced vibrations are also captured by the AprilTag reference Apriltag. During data processing, the reference signal is used to isolate and remove the UAV-induced vibration component from the structural response, yielding a more accurate estimate of the frame's true vibration. 

After takeoff and approaching the target frame, the UAV initiates an autonomous search procedure by rotating about the $z$-axis, as shown in Fig.~\ref{fig_phy_UAV_camera_view}. Once AprilTag is detected, the UAV adjusts its position, camera orientation, and stand-off distance to align with the marker correctly. The system then enters a tracking mode, continuously updating the UAV pose to maintain visual alignment with the AprilTag on the frame. As a result, the marker remains centered in the camera’s field of view throughout the recording process, even in the presence of minor UAV drift, eliminating the need for manual intervention.

\subsection{Data Processing and Analysis} We process videos using Motion Tracker Beta~\cite{FLOCH2023101424}, an open-source GUI-based motion tracking software. Table~\ref{tab_test_cases} lists the eight experimental cases tested under healthy (H) and damaged (D) conditions using different camera systems. The Galaxy S21 Ultra 5G and iPhone 15 Pro operate in UHD (3840 × 2160) at 30 and 60 frames per second (FPS), while the USB camera was tested at 1080p/60 FPS and 720p/120 FPS. Healthy cases correspond to the baseline prototype, while damaged cases are simulated by a reduction in natural frequency achieved by attaching an additional proof mass of approximately $240$ g at the center of the roof diaphragm. This approach is used to introduce a detectable frequency shift and streamline the experimental process rather than to change the stiffness of the column. Mass modification offers a similar decrease in natural frequency in a controlled and reversible way, whereas structural damage, such as cracking, usually shows up as stiffness degradation. As a trustworthy stand-in for assessing the platform's capacity to identify dynamic property changes suggestive of damage, this enables the evaluation of the detection system's sensitivity to frequency shifts without permanently altering the prototype.

%\vspace{0.5cm}

\begin{table}[!htbp]
\centering
\caption{Six healthy and damaged cases tested using different cameras and our UAV}
\label{tab_test_cases}
\vspace{-1mm}
{\scriptsize
\setlength{\tabcolsep}{5.5pt}
\begin{tabular}{cccccc}
\hline 
\begin{tabular}[c]{@{}c@{}}Camera\end{tabular}
& \multicolumn{2}{c}{\begin{tabular}[c]{@{}c@{}}Resolution\end{tabular}}
& \begin{tabular}[c]{@{}c@{}}FPS\end{tabular}
& \multicolumn{2}{c}{\begin{tabular}[c]{@{}c@{}}Case\end{tabular}}
\\
\hline
\multicolumn{4}{c}{\begin{tabular}[c]{@{}c@{}} \end{tabular}}
& \begin{tabular}[c]{@{}c@{}}Healthy\end{tabular}
& \begin{tabular}[c]{@{}c@{}}Damaged\end{tabular}
\\
\hline
\begin{tabular}[c]{@{}c@{}}Galaxy S21 Ultra\\5G\end{tabular}
& \begin{tabular}[c]{@{}c@{}}UHD (4K)\\UHD (4K)\end{tabular}
& \begin{tabular}[c]{@{}c@{}}3840$\times$ 2160\\3840$\times$2160\end{tabular}
& \begin{tabular}[c]{@{}c@{}}30\\60\end{tabular}
& \begin{tabular}[c]{@{}c@{}}HSC1\\HSC2\end{tabular}
& \begin{tabular}[c]{@{}c@{}}DSC1\\DSC2\end{tabular}
\\
\hline
\begin{tabular}[c]{@{}c@{}}iPhone 15 Pro\end{tabular}
& \begin{tabular}[c]{@{}c@{}}UHD (4K)\\UHD (4K)\end{tabular}
& \begin{tabular}[c]{@{}c@{}}3840$\times$2160\\3840$\times$2160\end{tabular}
& \begin{tabular}[c]{@{}c@{}}30\\60\end{tabular}
& \begin{tabular}[c]{@{}c@{}}HIC1\\HIC2\end{tabular}
& \begin{tabular}[c]{@{}c@{}}DIC1\\DIC2\end{tabular}
\\
\hline
\begin{tabular}[c]{@{}c@{}}USB Camera\\(IEights)\end{tabular}
& \begin{tabular}[c]{@{}c@{}}1080p\\720p\end{tabular}
& \begin{tabular}[c]{@{}c@{}}1920$\times$1080\\1280$\times$ 720\end{tabular}
& \begin{tabular}[c]{@{}c@{}}60\\120\end{tabular}
& \begin{tabular}[c]{@{}c@{}}HUC1\\HUC2\end{tabular}
& \begin{tabular}[c]{@{}c@{}}DUC1\\DUC2\end{tabular}
\\
\hline
\begin{tabular}[c]{@{}c@{}}UAV\end{tabular}
& \begin{tabular}[c]{@{}c@{}}720p\\720p\end{tabular}
& \begin{tabular}[c]{@{}c@{}}1280$\times$720\\848$\times$720\end{tabular}
& \begin{tabular}[c]{@{}c@{}}30\\60\end{tabular}
& \begin{tabular}[c]{@{}c@{}}HAC1\\HAC2\end{tabular}
& \begin{tabular}[c]{@{}c@{}}DAC1\\DAC2\end{tabular}
\\
\hline

\end{tabular}}
\end{table}

To evaluate the accuracy of the proposed vibration measurement approach, the natural frequencies obtained from different methods are compared. The comparison includes four cases: (1) the numerical prediction from the FE model for the healthy frame only. We cannot simulate damaged cases in the FE model, and it is unnecessary. (2) the frequencies measured by contact accelerometers in both healthy and damaged cases. (3) the frequencies identified from stand-off cameras using the Samsung Galaxy S21 Ultra, iPhone 15 Pro, and USB camera, and (4) the frequencies identified from our developed UAV.

\begin{comment}
    
% Maybe we could use this table (Max)
\begin{table}[!htbp]
\centering
\caption{Average Measured Frequencies and Standard Deviations for HAC and DAC Cases}
\vspace{-1mm}
{\scriptsize
\setlength{\tabcolsep}{6pt}
\begin{tabular}{cccc}
\hline 
\begin{tabular}[c]{@{}c@{}}Case~~\end{tabular}
& \begin{tabular}[c]{@{}c@{}}~Number of Tests~~\end{tabular}
& \begin{tabular}[c]{@{}c@{}}~Average Frequency~~\end{tabular}
& \begin{tabular}[c]{@{}c@{}}~Standard Deviation\end{tabular}
\\
\hline
\begin{tabular}[c]{@{}c@{}}HAC1\\HAC2\\DAC1\\DAC2\end{tabular}
& \begin{tabular}[c]{@{}c@{}}3\\3\\3\\3\end{tabular}
& \begin{tabular}[c]{@{}c@{}}5.38\\5.35\\4.75\\4.76\end{tabular}
& \begin{tabular}[c]{@{}c@{}}0.05\\0.01\\0.02\\0.02\end{tabular}
\\
\hline

\end{tabular}}
\end{table}
\end{comment}

\section{Results}
\label{results}

\subsection{Natural Modes and Frequencies}
The 3D FE model of the one-story frame building developed in COMSOL Multiphysics is used to identify its natural frequencies and corresponding mode shapes. The boundary conditions replicated the bolted base connections on the slip table, and an eigenfrequency analysis was carried out to extract the dominant vibration modes. The frequencies of the first three natural modes of the frame building modes are calculated as $5.2$ Hz, $16.0$ Hz, and $20.5$ Hz (see Table~\ref{table_percentage_error}). The first mode corresponds primarily to a sway in the longitudinal direction, while the higher modes involve combinations of lateral and torsional deformations. The first natural frequency serves as a reference for comparison with experimental results obtained from both contact- and non-contact-based measurements.

\subsection{Experimental Study}
\begin{figure}[htbp!]
    \centering
    \includegraphics[width=3.2in]{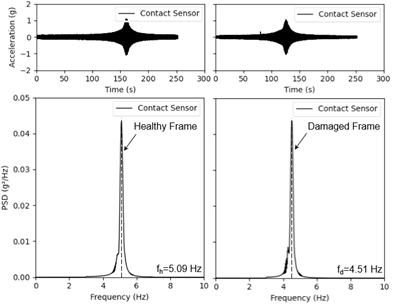}
    \vspace{-2mm}
    \caption{Time history of roof accelerations and the corresponding PSDs measured by a contact sensor.}
    \label{fig_roof_Contact_Sensor}
\end{figure}

Fig.~\ref{fig_roof_Contact_Sensor} shows the acceleration time histories and their corresponding power spectral densities (PSDs) obtained from the contact sensors for the healthy and damaged frame building. The healthy prototype shows a fundamental frequency of $5.09$ Hz, while the damaged prototype, simulated by the added proof mass, shows a reduced frequency of $4.51$ Hz. This shift demonstrates the sensitivity of contact sensors in detecting changes in structural dynamic properties and serves as the baseline reference for comparison with camera-based measurements. In addition, the fundamental frequency of the healthy frame is very close to the natural frequency of the first-mode predicted by the FE model in $f=5.2$ Hz.

% table was just added (Max)
\begin{table}[!htbp]
\centering
\caption{Percentage error relative to accelerometer reference values}
\vspace{-1mm}
\label{table_percentage_error}

{\scriptsize
\setlength{\tabcolsep}{2.0pt}
\begin{tabular}{cccccccc}
\hline
\multicolumn{2}{c}{\begin{tabular}[c]{@{}c@{}}Vibration\\ Measurement\end{tabular}} 
& \multicolumn{2}{c}{\begin{tabular}[c]{@{}c@{}}$f_h$=Frequency\\ Healthy (Hz)\end{tabular}} 
& \begin{tabular}[c]{@{}c@{}}Error(\%)= \\ $100\times$ \\ $\left| \frac{f_h - f_s}{f_s} \right|$\end{tabular} 
& \multicolumn{2}{c}{\begin{tabular}[c]{@{}c@{}}$f_d$=Frequency \\ Damaged (Hz)\end{tabular}}
& \begin{tabular}[c]{@{}c@{}}Error(\%)= \\ $100\times$ \\ $\left| \frac{f_h - f_s}{f_s} \right|$\end{tabular}
\\
\hline
\multicolumn{2}{c}{\begin{tabular}[c]{@{}c@{}}FE Model ($f_m$)\end{tabular}}
& \begin{tabular}[c]{@{}c@{}}-\end{tabular}
& \begin{tabular}[c]{@{}c@{}}5.20\end{tabular}
& \begin{tabular}[c]{@{}c@{}}- \end{tabular}
& \begin{tabular}[c]{@{}c@{}}-\end{tabular}
& \begin{tabular}[c]{@{}c@{}} - \end{tabular}
& \begin{tabular}[c]{@{}c@{}}- \end{tabular}
\\

\hline
\multicolumn{2}{c}{\begin{tabular}[c]{@{}c@{}}Contact Sensor ($f_s$)\end{tabular}}
& \begin{tabular}[c]{@{}c@{}}- \end{tabular}
& \begin{tabular}[c]{@{}c@{}}5.09\end{tabular}
& \begin{tabular}[c]{@{}c@{}}-\end{tabular}
& \begin{tabular}[c]{@{}c@{}}-\end{tabular}
& \begin{tabular}[c]{@{}c@{}}4.51\end{tabular}
& \begin{tabular}[c]{@{}c@{}}-\end{tabular}
\\

\hline
\begin{tabular}[c]{@{}c@{}}Forced \\Vibration\end{tabular}
& \begin{tabular}[c]{@{}c@{}}Non-\\Contact\\Sensor\end{tabular}
& \begin{tabular}[c]{@{}c@{}}HSC1\\HSC2\\HIC1\\HIC2\\HUC1\\HUC2\end{tabular}
& \begin{tabular}[c]{@{}c@{}}5.10\\5.12\\5.10\\5.10\\5.01\\5.00\end{tabular}
& \begin{tabular}[c]{@{}c@{}}0.2\\0.6\\0.2\\0.2\\1.6\\1.8\end{tabular}
& \begin{tabular}[c]{@{}c@{}}DSC1\\DSC2\\DIC1\\DIC2\\DUC1\\DUC2\end{tabular}
& \begin{tabular}[c]{@{}c@{}}4.51\\4.52\\4.51\\4.52\\4.43\\4.45 \end{tabular}
& \begin{tabular}[c]{@{}c@{}}0.0\\0.2\\0.0\\0.2\\1.8\\1.3\end{tabular}
\\
\hline
\begin{tabular}[c]{@{}c@{}}Free\\Vibration\end{tabular}
& \begin{tabular}[c]{@{}c@{}}UAV\\Camera\end{tabular}
& \begin{tabular}[c]{@{}c@{}}HAC1\\HAC2\end{tabular}
& \begin{tabular}[c]{@{}c@{}}5.38$\pm$0.05\\5.35$\pm$0.01\end{tabular}
& \begin{tabular}[c]{@{}c@{}}5.7\\5.1\end{tabular}
& \begin{tabular}[c]{@{}c@{}}DAC1\\DAC2\end{tabular}
& \begin{tabular}[c]{@{}c@{}}4.75$\pm$0.02\\4.76$\pm$0.02\end{tabular}
& \begin{tabular}[c]{@{}c@{}}5.3\\5.5\end{tabular}
\\
\hline

%\multicolumn{2}{c}{\begin{tabular}[c]{@{}c@{}}Contact Sensor ($f_s$)\end{tabular}}
%& \begin{tabular}[c]{@{}c@{}}- \end{tabular}
%& \begin{tabular}[c]{@{}c@{}}5.09\end{tabular}
%& \begin{tabular}[c]{@{}c@{}}-\end{tabular}
%& \begin{tabular}[c]{@{}c@{}}-\end{tabular}
%& \begin{tabular}[c]{@{}c@{}}4.51\end{tabular}
%& \begin{tabular}[c]{@{}c@{}}-\end{tabular}
%\\
%\hline
%\multicolumn{2}{c}{\begin{tabular}[c]{@{}c@{}}FE Model ($f_m$)\end{tabular}}
%& \begin{tabular}[c]{@{}c@{}}-\end{tabular}
%& \begin{tabular}[c]{@{}c@{}}5.20\end{tabular}
%& \begin{tabular}[c]{@{}c@{}}- \end{tabular}
%& \begin{tabular}[c]{@{}c@{}}-\end{tabular}
%& \begin{tabular}[c]{@{}c@{}} - \end{tabular}
%& \begin{tabular}[c]{@{}c@{}}- \end{tabular}
%\\
%\hline

\end{tabular}}
\end{table}

\begin{figure}[htbp!]
    \centering
    \includegraphics[width=3.2in]{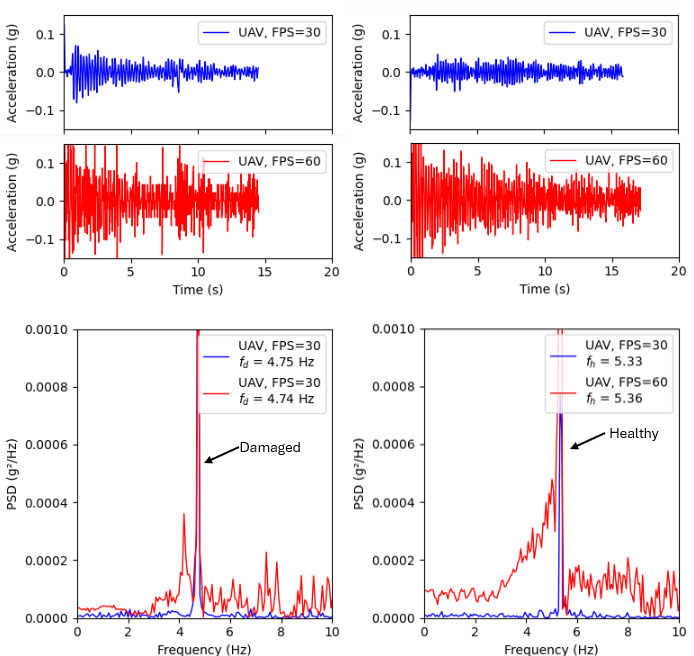}
    \vspace{-2mm}
    \caption{Time history of roof accelerations and the corresponding PSDs, measured by UAV.}
    \label{fig_roof_acc_UAV}
\end{figure}

Fig.~\ref{fig_roof_acc_UAV} shows the results obtained from the UAV camera at two frame rates, 30 and 60, as an example. The videos recorded by the UAV's camera and the stand-off cameras were processed to obtain the marker’s displacement. A region of interest around the roof marker was defined in the first frame and tracked across frames in motion-tracking software. The resulting pixel-coordinate trajectory was converted to physical displacement using a pixel-to-length calibration, and then acceleration was calculated using noise-robust numerical differentiation implemented using “PyNumDiff” in the software. Finally, the acceleration signal was transformed to the frequency domain, and the fundamental frequency was identified as the dominant peak in the PSD. For Galaxy S21, the natural frequencies are calculated as 5.10 Hz (30 FPS) and 5.12 Hz (60 FPS) for the healthy frame, and 4.51 Hz (30 FPS) and 4.52 Hz (60 FPS) for the damaged frame.

The results of all test cases for the different cameras and the UAV, along with the reference measurements from the contact sensor and the FE model, are summarized in Table~\ref{table_percentage_error}. Although the time histories obtained from the UAV camera do not match those recorded by the contact sensor shown in Fig.~\ref{fig_roof_acc_UAV}, particularly in terms of amplitude for FPS=30, the frequency-domain analysis remains highly reliable for both FPS = 30 and 60. The PSDs of the non-contact-based signals accurately capture the fundamental frequencies and their shifts, showing close agreement with the contact sensor measurements despite the motion of the UAV. 

 The FE model was developed specifically to validate the baseline dynamic characteristics of the prototype; therefore, only the healthy state was simulated numerically, and we have no damaged frequency and errors. Since the damaged condition was introduced as a temporary experimental modification via an added proof mass, the contact sensor measurements served as the primary reference (ground truth) without errors. It should be noted that the FE model represents only the undamaged (healthy) structural condition and does not account for damage scenarios, which limits its use for direct validation of damaged-state responses.  
 \begin{comment}

I added this part "It should be noted that the FE model represents only the undamaged (healthy) structural condition and does not account for damage scenarios, which limits its use for direct validation of damaged-state responses. "
\end{comment}  

For the iPhone 15 Pro, the measured natural frequencies were 5.10 Hz (HIC1, 30 FPS) and 5.10 Hz (HIC2, 60 FPS) for the healthy frame and 4.51 Hz (DIC1, 30 FPS) and 4.52 Hz (DIC2, 60 FPS) for the damaged frame. These values are in excellent agreement with the contact sensor measurements, demonstrating that the iPhone provides consistent frequency identification across both frame rates. In addition, the corresponding percent error relative to the contact-sensor reference shows that the smartphone-based cases (HIC/HSC and DIC/DSC) achieve errors of only 0 to 0.6$\%$ under healthy and damaged conditions. The results also confirm that the non-contact method can reliably capture the frequency shift caused by the added proof mass. The USB camera showed slightly larger deviations compared to the smartphone results. For the healthy frame, the measured frequencies were $5.01$ Hz (HUC1, 60 FPS) and $5.0$ Hz (HUC2, 120 FPS), while for the damaged frame, the frequencies were $4.43$ Hz (DUC1, 60 FPS) and $4.45$ Hz (DUC2, 120 FPS). Although the USB camera still captured the reduction in frequency due to damage, the absolute values were marginally lower than those of the contact sensors. This variation is likely due to its lower resolution, which affects the accuracy of displacement tracking. Consistent with this observation, the cases with USB-camera show slightly higher errors of about $1.3$ to $1.8\%$. Nevertheless, the results demonstrate that higher frame rates (120 FPS) help to improve temporal accuracy, making the USB camera a viable low-cost option for vibration-based damage detection.

In contrast to stationary cameras, UAV-based measurements exhibited larger deviations from the reference data. As shown in Table~\ref{table_percentage_error}, the UAV identified natural frequencies of $5.38$ Hz (30 FPS) and $5.35$ Hz (60 FPS) under healthy conditions and $4.75$ Hz and $4.76$ Hz under damaged conditions. The reported standard deviations were calculated from three experimental trials. These measurements correspond to percent errors ranging from $5.1\%$ to $5.7\%$ relative to the contact sensor measurements.
The reduced accuracy can be attributed to several platform-specific factors. First, rotor-induced vibrations introduce mechanical noise that degrades the quality of the recorded signal. Second, the autonomous AprilTag-based positioning system maintains alignment within a tolerance band rather than enforcing an exact lock, as continuous fine-scale corrections would require constant UAV repositioning and would create additional video instability and measurement noise. Third, the relatively short free-vibration recording limits the effective signal window for analysis, resulting in broader spectral peaks and lower precision in frequency estimation. Finally, the lower resolution of the onboard camera reduces tracking accuracy compared with the 4K smartphones used in the stationary setup.

The observed error range of $5.1-5.7\%$ is similar to vibration measurement errors reported in the literature for high-end systems. For example, investigations using commercial systems have shown comparable issues with motion blur and stabilityn ~\cite{ReliabilityUAV2023, CrossCorrUAV2017}. This implies that our platform achieves an adequate level of accuracy for identifying notable frequency shifts, even though it costs only $10-20\%$ of a commercial inspection drone (e.g., DJI Matrice series).

Despite these absolute errors, the UAV platform successfully captured the fundamental frequency shift. Specifically, the measured decrease of approximately $0.6$ Hz between healthy and damaged states closely matches the trend observed from the contact sensors, demonstrating the ability of the low-cost aerial platform to detect structural changes through vibration-based analysis.

\section{Conclusions}
\label{conclusion}

This paper demonstrates the use of motion tracking for non-contact vibration measurement and damage detection on a frame building in a laboratory environment. The results obtained from smartphones and a USB camera were in close agreement with the predictions of the contact sensor and the FE model, with errors below $1.8\%$, validating their reliability as stationary monitoring tools.

Regarding the UAV platform, the experimental results highlight a distinct trade-off between cost-effectiveness and absolute measurement precision. The UAV measurements exhibited a higher percentage error, ranging from ($5.1 - 5.7\%$), compared to stationary devices (contact sensors, cellphones, and USB cameras). This deviation is primarily attributable to the inherent instability of the hovering platforms, where rotor-induced vibration and drift affected the displacement signals. However, the critical finding is that the UAV successfully identifies the structural damage. Despite the higher absolute error, the system clearly captured the shift in fundamental frequency caused by the added mass. This confirms that low-cost UAVs are sufficiently reliable for detecting significant trends of structural degradation as stationary devices. Compared to other research using expensive commercial UAVs, our custom-built UAV is cost-effective and sustainable. Rather than acquiring high-precision absolute displacement measurements, its inspection performance satisfies the requirements for preliminary structural health monitoring, where the main goal is to discover substantial shifts in natural frequencies suggestive of damage.

This study represents an initial step toward applying low-cost UAV platforms to non-contact vibration-based damage detection. The proposed approach was validated using a single-story laboratory prototype subjected to controlled shaker excitation. In the “damage” case, structural damage was simulated by adding a proof mass to the roof to induce a shift in natural frequency, rather than introducing stiffness degradation (e.g., cracking or connection loosening), which can also lead to frequency reduction.

%In addition, the UAV experiments were conducted in a controlled indoor laboratory environment. 
In practical field applications, environmental disturbances, such as wind, can affect flight stability and measurement quality. Therefore, further studies are required to evaluate the performance of the UAV-based system on larger-scale structures under field or ambient conditions and with more realistic damage scenarios.

To further improve inspection accuracy and robustness, future work will explore the deployment of multiple cooperative UAV swarms. Such a system could inspect civil structures from different points of view, potentially improving measurement quality and spatial coverage. Due to the relatively low cost of individual UAV platforms, a multi-UAV system can also provide enhanced robustness and scalability for large-scale structural health monitoring applications.


%%%%%%%%%%%%%%%%%%%%%%%%%%%%%%%%%%%%%%%%%%%%%%%%%%%%%%%%%%%%%%%%%%%%%%%%%%%%%%%%

%%%%%%%%%%%%%%%%%%%%%%%%%%%%%%%%%%%%%%%%%%%%%%%%%%%%%%%%%%%%%%%%%%%%%%%%%%%%%%%%

%%%%%%%%%%%%%%%%%%%%%%%%%%%%%%%%%%%%%%%%%%%%%%%%%%%%%%%%%%%%%%%%%%%%%%%%%%%%%%%%
%\section*{APPENDIX}


\section*{ACKNOWLEDGMENT}

The authors acknowledge the financial support provided by the NSF Expand AI ARISE project (No. 2434916), the NSF CREST Center for Multidisciplinary Research Excellence in Cyber-Physical Infrastructure Systems (MECIS) (No. 2112650), and the NSF MSI program (No. 2318682).

%This work is supported by the SLA (Scientific Leadership Award) program through DHS (Department of Homeland Security) Award No. 21STSLA00009-01-00. The authors also acknowledge the partial funding provided by the CREST MECIS program through NSF (National Science Foundation) Award No. 2112650 and the MSI program through NSF Award No. 2318682.

%%%%%%%%%%%%%%%%%%%%%%%%%%%%%%%%%%%%%%%%%%%%%%%%%%%%%%%%%%%%%%%%%%%%%%%%%%%%%%%%

%References are important to the reader; therefore, each citation must be complete and correct. If at all possible, references should be commonly available publications.

\bibliographystyle{IEEEtran}
\bibliography{BibFiles/IEEEabrv, BibFiles/main_ref}

\end{document}